\documentclass{article}

\usepackage[nonatbib,preprint]{neurips_2022}

\usepackage[utf8]{inputenc} 
\usepackage[T1]{fontenc}    
\usepackage{url}            
\usepackage{booktabs}       
\usepackage{amsfonts}       
\usepackage{nicefrac}       
\usepackage{microtype}      
\usepackage{xcolor}         
\usepackage{amsmath,amssymb}
\usepackage{cleveref}       
\usepackage{graphicx}
\usepackage{tabularx}
\usepackage{tikz}
\usepackage[backend=biber]{biblatex}
\usepackage{caption}
\usepackage{subcaption}
\usepackage{wrapfig}
\usepackage{lipsum}
\usepackage{comment}
\usepackage[symbol]{footmisc}

\usetikzlibrary{shapes.geometric, arrows}

\tikzstyle{node} = [rectangle, ,text centered, draw=black, inner sep=0.75em]
\tikzstyle{decoder} = [trapezium, shape border rotate=90, trapezium angle=77.5, minimum height=0.8cm,text centered, draw=black, inner sep=0.75em]
\tikzstyle{arrow} = [->,>=stealth]

\title{Deep Active Latent Surfaces for Medical Geometries}

\newcommand{\dasm}{{\it DASM}}
\newcommand{\dasr}{{\it DASM+R}}
\newcommand{\dsdf}{{\it DeepSDF}}
\newcommand{\dsds}{{\it SIREN+DeepSDF}}
\newcommand{\dmlp}{{\it DUAL-MLP}}
\newcommand{\ours}{{\it DALS}}
\newlength{\snyd}

\newcommand{\norm}[1]{\left\lVert#1\right\rVert}
\newcommand{\R}{\ensuremath{\mathbb{R}}}

\newcommand{\cD}{\mathcal{D}}
\newcommand{\cF}{\mathcal{F}}
\newcommand{\cL}{\mathcal{L}}
\newcommand{\cM}{\mathcal{M}}

\newcommand{\bx}{\mathbf{x}}
\newcommand{\bz}{\mathbf{z}}

\newcommand{\bZ}{\mathbf{Z}}

\DeclareMathOperator*{\argmin}{\arg\min}

\newcommand{\note}[1]{{\color{red} #1}}

\author{%
  Patrick~M. Jensen\\
  \texttt{patmjen@dtu.dk}
  \And
  Udaranga Wickramasinghe\\
  \texttt{udaranga.wickramasinghe@epfl.ch}
  \And
  Anders~B. Dahl\\
  \texttt{abda@dtu.dk}
  \And
  Pascal Fua$^*$\\
  \texttt{pascal.fua@epfl.ch}
  \And
  Vedrana~A. Dahl\footnote{}\\
  \texttt{vand@dtu.dk}
}

\crefname{figure}{Fig.}{Fig.}
\crefname{table}{Tab.}{Tab.}

\begin{document}

\maketitle


\footnotetext{Equal supervision.}
\begin{abstract}

Shape priors have long been known to be effective when reconstructing 3D shapes from noisy or incomplete data.  When using a deep-learning based shape representation, this often involves learning a latent representation, which can be either in the form of a single global vector or of multiple local ones. The latter allows more flexibility but is prone to overfitting. 
In this paper, we advocate a hybrid approach representing shapes in terms of 3D meshes with a separate latent vector at each vertex. During training the latent vectors are constrained to have the same value, which avoids overfitting. For inference, the latent vectors are updated independently while imposing spatial regularization constraints. We show that this gives us both flexibility and generalization capabilities, which we demonstrate on several medical image processing tasks. 

\end{abstract}


\section{Introduction}


3D shape reconstruction from noisy or incomplete data usually benefits from the judicious use of shape priors. When using a deep-learning based shape representation, this usually means 
searching for an appropriate latent representation under regularization losses.
This latent vector representation can take the form of either a single global latent vector per shape or a grid of latent vectors. 

In either case, it is a challenge to balance regularization against quality of fit to the data. With too much regularization the recovered shapes are too smooth and fine details are lost. With too little, robustness to noise and generalization capabilities will suffer. Our insight is that, when using multiple latent vectors, instead of imposing regularity constraints directly on the surface, we can impose them on the latent vectors. This is effective because, if a latent vector has been trained to model a sharp feature, requiring that this vector be similar to its neighbors will not detract from that. To this end, we represent surfaces as triangulated meshes and advocate using a separate latent vector at each vertex while imposing smoothness constraints on these vectors. To exploit this, we borrow the regularization idea from early approaches to 3D shape modeling~\cite{kass1988snakes,terzopoulos1988symmetry,terzopoulos1988constraints} that predate deep learning and apply it to the latent vectors instead of the vertex positions.

\begin{figure}[ht]
    \centering
    \includegraphics[width=\linewidth]{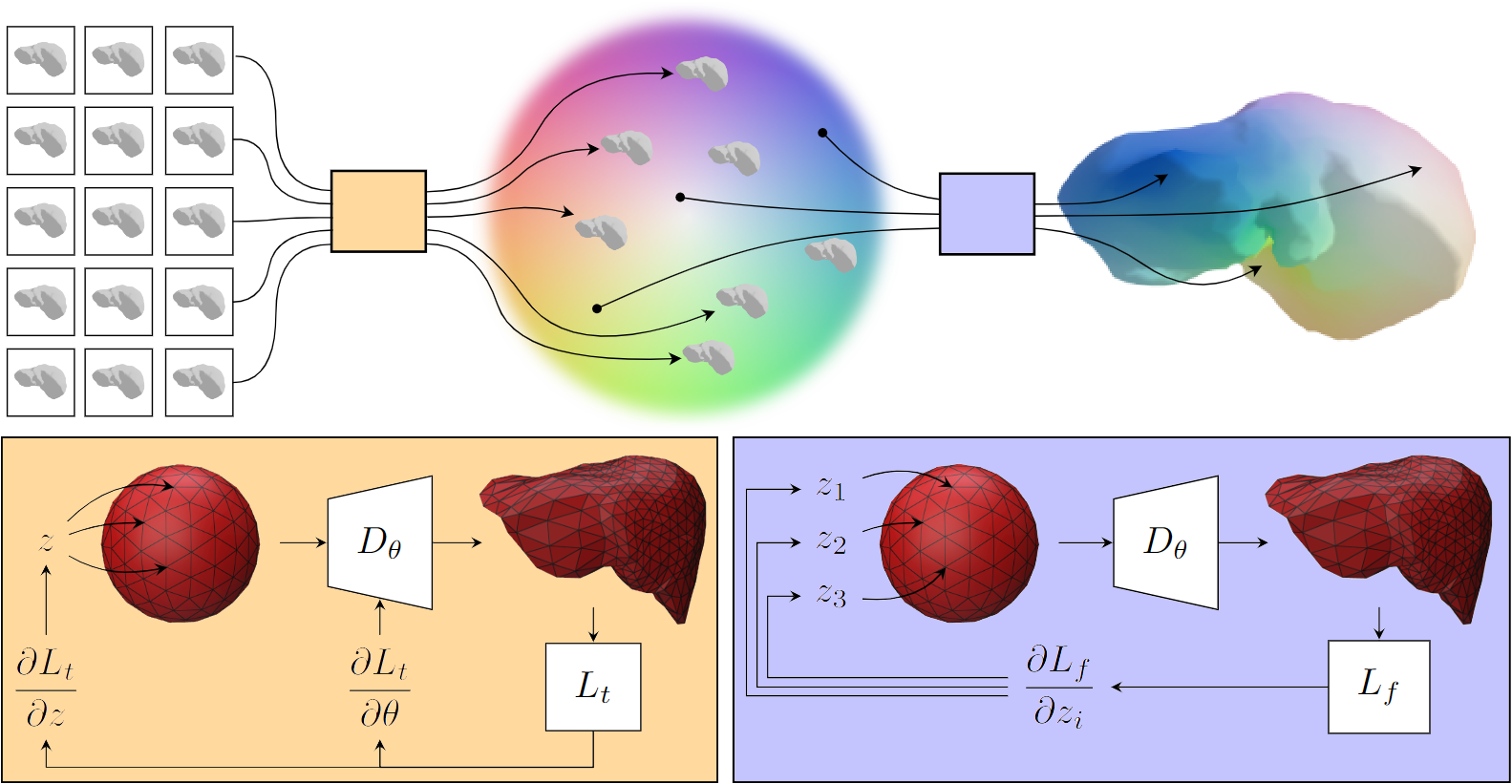}
    \caption{{\bf \ours{} overview.}  {\bf (top)} Given a set of training shapes, we use an auto-decoding approach to learning a latent space of shapes. {\bf (bottom left)} To each training shape is associated a single latent vector $\mathbf{z}$ that is used to compute a translation for each vertex of sphere to minimize the distance between the deformed sphere and the training shape. {\bf (bottom right)} At inference time, the latent vectors $\mathbf{z_1},\mathbf{z_2},\ldots,\mathbf{z_N}$ at each one the $N$ sphere vertices are allowed to change independently to minimize a weighted sum of a data loss function and a regularization term that prevents neighboring latent vectors from being too different from each other.}
    \label{fig:method-overview}
\end{figure}

More specifically, at training time, we use an auto-decoding approach~\cite{park2019deepsdf} to jointly learn the network weights along with one {\it common} latent vector for all vertices of each training sample. By contrast, at inference time, we allow the latent vectors to be different at each vertex while enforcing consistency of the vectors by minimizing a regularization loss. Fig.~\ref{fig:method-overview} depicts our Deep Active Latent Surfaces (DALS) approach.  It makes our model both easy to train from relatively small datasets and very expressive: Because, we learn a single latent vector per shape, we do not require the training set to be huge. At model fitting time, we do not need to find an individual vector that accurately represents a {\it whole} shape, which can be difficult when the shape is complex. Instead, we can smoothly blend a number of vectors to model such a shape, while preserving sharp features. Furthermore, explicitly using a triangulated mesh instead of a grid of latent vectors makes our model more compact and simpler to train because it does not have to waste capacity on modeling empty regions.


\section{Related Work}

\paragraph{Active Surface Models}

Active contour models are used to refine contours according to local image properties while remaining smooth. They were first introduced in~\cite{kass1988snakes} for interactive delineation and then extended for many different purposes~\cite{fua1996model}. Active surface models operate on the same principle~\cite{terzopoulos1988constraints,terzopoulos1988symmetry} but replace the contours by triangulated meshes to model 3D surfaces. They have proved very successful for medical~\cite{he2008comparative,mcinerney1995dynamic} and cartographic applications~\cite{fua1995object}, among others,  and are still being improved~\cite{jorstad2015neuromorph,leventon2000stastical,prevost2013incorporating}. In recent years, Deep Neural Networks (DNNs) have been used to evaluate the energy that the active contours minimize~\cite{hatamizadeh2020endtoend,marcos2018learning} and, in \cite{liang2020polytransform,ling2019fast,peng2020deep}, they are used to directly predict vertex offsets. In~\cite{wickramasinghe2021deep}, active surface models are embedded in special purpose network layers that regularize surface meshes using the same semi-implicit scheme as the original active contours~\cite{kass1988snakes}.

Correctly balancing the relative influence of the data and regularization terms to avoid over-smoothing while being robust to noise remains a challenge for all these approaches.  In \cite{liang2020polytransform} smoothing is only added as a loss during training but not during inference. In \cite{wickramasinghe2021deep}, smoothing is made adaptive to allow sharp edges. Recently, it has been proposed to replace smoothing with preconditioned gradient descent of the external energy~\cite{nicolet2021large}.  In all these approaches, the regularization  tends to flatten sharp geometric features of the mesh geometry, which almost always results in  over-smoothing, despite the goodness of fit at a global level. In our work, we side step these issues by focusing the smoothing on the latent space. 

\paragraph{Neural Shape Modeling}

Deep-learning methods are now routinely used to model 3D shapes. Most methods rely on auto-encoders or auto-decoders to produce latent vectors that parameterize the target shapes in terms of triangulated meshes~\cite{hanocka2020point2mesh,litany2018deformable,morreale2022neural}, tetrahedral meshes \cite{gao2020learning,shen2021dmtet}, surface patches \cite{groueix2018papier}, point clouds~\cite{achlioptas2018learning,peng2021shape}, voxel grids~\cite{brock2016generative,dai2017shape}, occupancy functions \cite{chen2019learning,mescheder2019occupancy,peng2020convolutional}, signed and unsigned distance fields \cite{chibane2020implicit,juhl2021implicit}, and neural splines~\cite{williams2021nkf}. 

These methods can be classified as those that use a single latent vector to represent a complete shape and those that use multiple ones. Among those that use a single one are the approaches of~\cite{chen2019learning,juhl2021implicit,litany2018deformable,park2019deepsdf}. They are effective but accurately representing all the details may require more than one latent vector as in the methods of~\cite{chibane2020implicit,jiang2020local,mehta2021modulated,peng2020convolutional} that rely on a grid of latent vectors and a shared decoder to represent the signed distance function of a complete shape. Since each latent vector only has to describe a small part of the complete shape, this greatly increases the model flexibility. It also allows for a much smaller decoder network, which makes inference faster. However, storing a full grid of latent vectors means that some latent vectors are wasted on representing empty space. This increases memory use and training time. The approach of~\cite{chabra2020deep,liu2020neural}  avoids this by using a sparse grid from which unused grid cells are removed. In~\cite{martel2021acorn,takikawa2021nglod}, a tree structures is used to construct sparse multiscale representations. However, these methods either require prior knowledge about the surface or periodic updates of the sparse data structure as it deforms, which makes training more complex. In \cite{muller2022instant}, these issues are alleviated by using a spatial hash encoding which allows the model to implicitly allocate more capacity to regions near the surface. However, this method is designed to represent single shapes and would require non-trivial extensions for model fitting purposes. In our work, we also rely on multiple latent vectors but require neither {\it a priori} knowledge about the surface nor complex adaptation of a data structure. 




\section{Deep Active Latent Surfaces}

We now describe our Deep Active Latent Surface (\ours{}) approach, which is illustrated by \cref{fig:method-overview}. We represent watertight 3D shapes by triangulated spheres with a latent vector at each vertex. This latent vector along with the vertex coordinates is fed to a decoder $\mathcal{D}_\theta$ that generates an offset vector that is then used to translate the vertex to its final position. Once all the vertices have been translated, we have the final shape such as the one shown on the top right part of \cref{fig:method-overview}.  At training time, we use the same latent vector for all vertices whereas, at model fitting time, we allow them to be different but impose spatial consistency on the vectors. This does not preclude the modeling of sharp features because such features can be predicted by individual latent vectors. 

\subsection{Training Scheme}\label{sec:training}

 Formally, let $\cD_{\Theta}$ be a neural network with weights $\Theta$ that takes as input a $d$-dimensional latent vector $\bz$ and a 3D location $\bx$ and returns an offset $\cD_{\Theta}(\bz,\bx)$. Given a triangulated sphere with $V$ vertices and $F$ facets, we denote by $\cM_{\theta}(\bz)$ the deformed mesh we obtain by translating each vertex $\bx_v$ by  $\cD_{\Theta}(\bz,\bx_v)$ for all $v$ between 1 and $V$.
 
 Let us further assume we are given a set of $N$ training shapes $S = \{ S_1,\ldots,S_N\}$. As in~\cite{park2019deepsdf}, we can simultaneously learn $\Theta$ and a $\bz_i$ for each $S_i$ by looking for 
 \begin{align}
 \Theta^*,\bz_1^*,\ldots,\bz_N^* &=  \argmin_{\Theta,\bz_1,\ldots,\bz_N} \sum_{i=1}^N \cL_\text{dat}(\Theta,\bz_i,S_i) \; , \label{eq:autoF} \\
 \cL_\text{dat}(\Theta,\bz,S_i) &=  L_\text{cf}(\cM_{\theta}(\bz),S_i) + \lambda_\text{reg} L_\text{reg}(\cM_{\theta}(\bz))  + \lambda_\text{n} \| \bz \|^2 \;  , \nonumber
\end{align}
where $L_\text{cf}$ is the Chamfer distance \cite{smith2019geometrics},  $L_\text{reg}$ is shape regularization term, and $\lambda_\text{reg}$ and $\lambda_\text{n}$ are weighting constants.

In practice, we take $\cD_{\Theta}$ to be an MLP with three hidden layers of size $724$, $724$, and $362$, which takes as input a concatenation of $\bx$ and $\bz$. We use ReLU activations for the hidden layers and none for the last layer. Before each ReLU activation we use layer normalization~\cite{ba2016layer}. Our initial spherical triangulation is a subdivided icosahedron. We use a pointwise MLP instead of a mesh based decoder because we want to learn a mapping from the surface of a sphere conditioned on a latent vector rather than a mapping from a specific template mesh. To this end, we also randomly rotate the template 
{
\parfillskip=0pt
\parskip=0pt
\par}
\begin{wrapfigure}{R}{0.5\linewidth}
	\centering
	\input{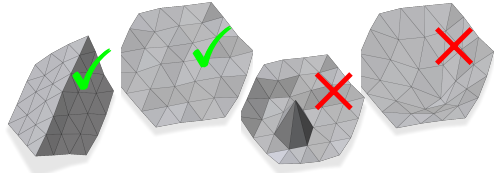}
	\vspace{-\baselineskip}
	\caption{{\bf Behavior of $L_\mathrm{reg}$.} Minimizing $L_\mathrm{reg}$ will leave the regular meshes on the left unchanged, even though the first one exhibits a sharp crease. In contrast, doing so with the meshes on the right, will increase the regularity of the triangles and smooth out the isolated outlier.}
	\label{fig:tri-quality-loss}
\end{wrapfigure}
during training to specific vertex placement. This enables us to use any template mesh without changing the decoder. We will take advantage of this at inference time by progressively increasing the mesh resolution.

The $L_\text{reg}$ term in Eq.~\ref{eq:autoF} is intended to encourage the generation of high quality meshes. We could have expressed it in terms of the Laplacian as is often done but we found experimentally that it tends to result in meshes that are too smooth. Instead, as in~\cite{bhatia1990two}, we take it to be 
\begin{equation}
    	L_\text{reg}(\mathcal{M}) = 1 - \frac{4\sqrt{3}}{|\mathcal{F}|} \sum_{f \in \mathcal{F}} \frac{A_f}{a_{f}^2 + b_{f}^2 + c_{f}^2} \; ,
    	\label{eq:smooth}
\end{equation}
where $\cF$ stands for the mesh facets, $a_f, b_f, c_f$ for the lengths of the three edges of facet $f$, and $A_f$ for its area. It is easy to compute and has favorable properties for numerical optimization \cite{shewchuk2002good}. In practice, minimizing $L_\text{reg}$ promotes regularity without directly penalizing high frequency features, as illustrated by \cref{fig:tri-quality-loss}.

\subsection{Fitting Scheme}
\label{sec:inference}

Let $\cD=\cD_{\Theta^*}$ be the network we trained in Section~\ref{sec:training} and let $\bZ \in \R^{V\times d}$  whose $V$ rows are latent vectors, one for each vertex of our triangulated sphere. We now denote by $\cM(\bZ)$ the mesh we obtain by shifting each vertex $\bx_v$ by $\cD(\bZ[v])$. In other words, we now assign to each vertex a \emph{different} latent vector and use it to compute the corresponding translation from the initial sphere to where it should be. To fit our model to data, we look for
\begin{equation}
\begin{aligned}
    \textbf{Z}^* =& \argmin_{\textbf{Z}} \cL_\text{task}(\cM(\bZ))\, +\\& \lambda_\text{reg} L_\text{reg}(\mathcal{M}(\bZ)) + \lambda_\text{dir} L_\text{dir}(\bZ) \; ,
    \label{eq:fullLoss}
\end{aligned}
\end{equation}
where $\cL_{\text{task}}$ is a task-specific loss function, $L_\text{reg}$ is the geometric regularization loss of Eq.~\ref{eq:smooth}, $L_\text{dir}$ is a new regularization term designed to enforce consistency of the latent vectors across the surface, and $\lambda_\text{reg}$ and $\lambda_\text{dir}$ are weighting constants. In practice, $\cL_{\text{task}}$ can be 
expressed in terms of Chamfer distance to fit points, slice Chamfer to fit curve annotations, and SDF gradients for segmentation purposes as discussed in Section~\ref{sec:experiments}. 

Inspired by active surfaces \cite{kass1988snakes,terzopoulos1988constraints,terzopoulos1988symmetry,wickramasinghe2021deep}, we use Dirichlet energy~\cite{nicolet2021large,solomon2014laplace} to define $L_\text{dir}$. We write 
\begin{equation}
    L_\text{dir}(\bZ) = \text{Tr}(\bZ^T\textbf{L}^p\bZ) \; ,
\end{equation}
where $\textbf{L}$ is the uniform Laplacian matrix and $p$ is an integer power. Note that $\nabla L_\text{dir}(\textbf{Z}) = \textbf{L}^p\bZ$, meaning that a gradient step corresponds to $p$ iterations of Laplacian smoothing of the latent vectors. We found $p=2$ to work well.

Weight $\lambda_\text{dir}$ controls how constrained the fitting is by the prior information contained in the latent space parameterization.  When $\lambda_\text{dir} \to \infty$, all latent vectors will have the same value and our approach reverts to a global approach with one single latent vector. For small values of $\lambda_\text{dir}$, the model becomes much more flexible as the values of the latent vector can more easily change from vertex to vertex. 

One could also allow independant training vectors during training. However, in our experience, this causes the decoder to produce severely self-intersecting meshes as it has too much freedom. 



\section{Experiments}
\label{sec:experiments}

We demonstrate the benefits of \ours{} on several medical image processing tasks. We train all models to learn a latent representation of livers and spleens using data from the Medical Segmentation Decathlon (MSD)~\cite{antonelli2021medical} (CC~BY-SA~4.0 license). To create ground-truth meshes, we first resampled the annotated images so that their voxel size is $1\times$1$\times1$ mm and then used marching cubes~\cite{lewiner2003efficient} to extract isosurfaces. We standardize the surfaces to have zero mean and be contained in the unit sphere.

The datasets contains 111 livers and 41 spleens. We train a model on the first 71 livers and hold out the last 40 for evaluation. We also train another model on the first 31 spleens with the last 10 held out for evaluation. We augment the training data using the recent PointWOLF algorithm~\cite{kim2021point} to create 100 new shapes for each training shape. PointWOLF applies a smoothly varying non-rigid transformation to mesh vertices and yields diverse and realistic augmentations.

We use 128 dimensional latent vectors and an icosahedron subdivided 3 times for training and 4 times for fitting as a template for the decoder.
To learn these vectors and the decoder weights we solve the minimization problem of Eq.~\ref{eq:autoF} with $\lambda_\mathrm{reg} = 10^{-4}$ and $\lambda_\mathrm{n} = 10^{-3}$. To this end, we use the ADAM optimizer \cite{kingma2014adam}. We set the learning rate to 0.002, the momentum terms to $\beta_1 = 0.9, \beta_2 = 0.999$ and train for 24 hours on a single NVIDIA Tesla V100 GPU (ca. 7,500 epochs). If the loss does not improve for 100 epochs we reduce the learning rate by a factor 2, down to a minimum of $10^{-5}$.

\paragraph{Baselines} 

We compare our model against the following baselines: \dsdf{} \cite{park2019deepsdf}, \dsds{} where the training and inference method is as in \dsdf{} but with a SIREN~\cite{sitzmann2020implicit} based decoder, and the \dmlp{} approach of~\cite{mehta2021modulated}. 
We also compare to \dasm{}, the active surfaces of~\cite{wickramasinghe2021deep},  along with an improved version that we dub \dasr{} because it adds a re-meshing step during fitting to avoid self-intersections. These are the only two baselines that do not rely on a learned shape prior and simply promote smoothness. 

As the \dmlp{} authors did not release code, we implemented two separate versions of it, one that uses one single latent vector per shape (global) and one that uses several (local). All these methods were trained as recommended in the relevant papers.

\subsection{Shape Reconstruction from 3D Point Clouds}
\label{sec:recon-livers}

\paragraph{Experimental Setup.}

We test the ability of the latent vector models to reconstruct unknown shapes from a given class, here the liver and the spleen, by randomly and uniformly sampling 2,500 points across the test surface and attempting to reconstruct from them by minimizing the loss of Eq.~\ref{eq:fullLoss}.  For \ours{}, \dasm{}, and \dasr{} that use a mesh-based representation, we take $\cL_\mathrm{task}$ to be the Chamfer distance. For the other methods, we take it to be the mean absolute SDF value at the sample points. For all methods, we use the ADAM optimizer to minimize their fitting losses. For this task, we set $\lambda_\mathrm{reg} = 0.001$ and $\lambda_\mathrm{dir} = 0.2$. Finally, for \dasr{} and \ours{} we post process the results using five iterations of Botsch-Kobbelt remeshing \cite{botsch2004remeshing}.

To evaluate the reconstructions, we use the Chamfer distance, the Hausdorff distance, and the F-score \cite{tatarchenko2019single,knapitsch2017tanks} at 1\% and 2\% of the surface's bounding sphere diameter. We also evaluate the mesh quality of the reconstructions using the quality measure of Eq.~\ref{eq:smooth} and the percentage of self-intersecting faces.

\paragraph{Results.}

\setlength{\snyd}{1.5\baselineskip}
\begin{figure}[h]
    \centering
    \newcommand{\insetwidth}{0.7\linewidth}
    \begin{subfigure}[b]{0.245\linewidth}
        \centering
        \includegraphics[width=\linewidth,trim={100px 0 140px 0},clip]{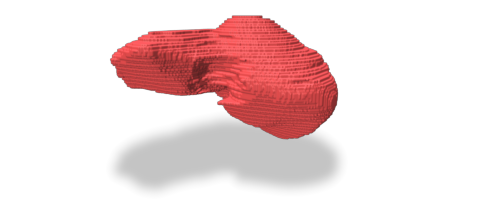}
        \vspace{-\snyd}
        \caption{Ground truth.}
    \end{subfigure}
    \begin{subfigure}[b]{0.245\linewidth}
        \centering
        \begin{tikzpicture}[every node/.style={anchor=south west,inner sep=0pt},x=1mm, y=1mm]   
             \node (fig1) at (0,0)
               {\includegraphics[width=\linewidth,trim={100px 0px 140px 0},clip]{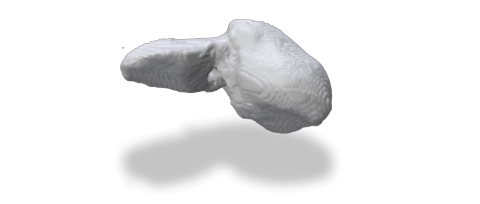}};
             \node (fig2) at (0,0.5)
               {\includegraphics[width=\insetwidth,trim={100px 50px 140px 0},clip]{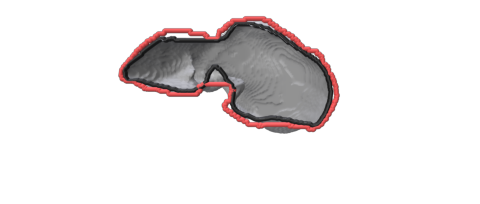}};  
        \end{tikzpicture}
        \vspace{-\snyd}
        \caption{\dsdf{}.}
    \end{subfigure}
    \begin{subfigure}[b]{0.245\linewidth}
        \centering
        \begin{tikzpicture}[every node/.style={anchor=south west,inner sep=0pt},x=1mm, y=1mm]   
             \node (fig1) at (0,0)
               {\includegraphics[width=\linewidth,trim={100px 0px 140px 0},clip]{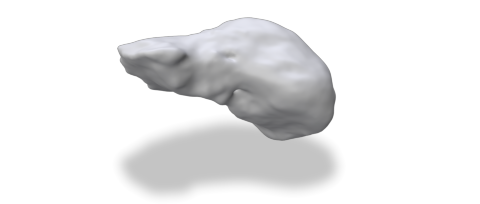}};
             \node (fig2) at (0,0.5)
               {\includegraphics[width=\insetwidth,trim={100px 50px 140px 0},clip]{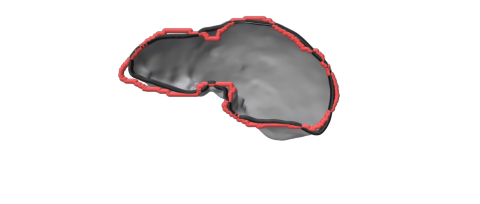}};  
        \end{tikzpicture}
        \vspace{-\snyd}
        \caption{\dsds{}.}
    \end{subfigure}
    \begin{subfigure}[b]{0.245\linewidth}
        \centering
        \begin{tikzpicture}[every node/.style={anchor=south west,inner sep=0pt},x=1mm, y=1mm]   
             \node (fig1) at (0,0)
               {\includegraphics[width=\linewidth,trim={100px 0px 140px 0},clip]{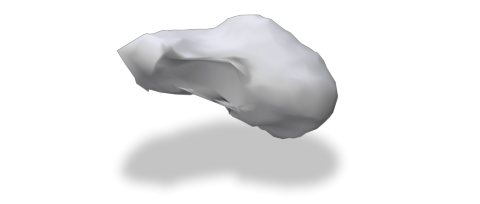}};
             \node (fig2) at (0,0.5)
               {\includegraphics[width=\insetwidth,trim={100px 50px 140px 0},clip]{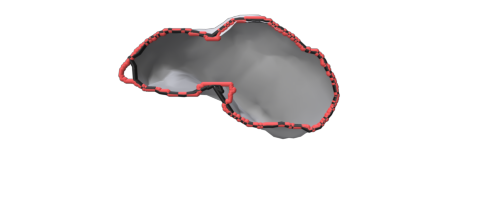}};  
        \end{tikzpicture}
        \vspace{-\snyd}
        \caption{\dasm{}.}
    \end{subfigure}
    \\
    \begin{subfigure}[b]{0.245\linewidth}
        \centering
        \begin{tikzpicture}[every node/.style={anchor=south west,inner sep=0pt},x=1mm, y=1mm]   
             \node (fig1) at (0,0)
               {\includegraphics[width=\linewidth,trim={100px 0px 140px 0},clip]{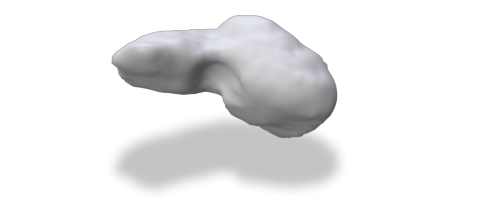}};
             \node (fig2) at (0,0.5)
               {\includegraphics[width=\insetwidth,trim={100px 50px 140px 0},clip]{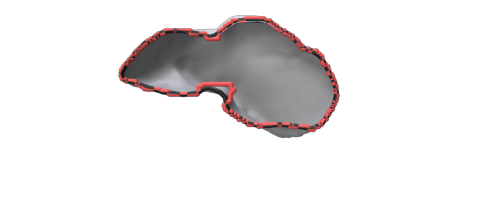}};  
        \end{tikzpicture}
        \vspace{-\snyd}
        \caption{\dasr{}.}
    \end{subfigure}
    \begin{subfigure}[b]{0.245\linewidth}
        \centering
        \begin{tikzpicture}[every node/.style={anchor=south west,inner sep=0pt},x=1mm, y=1mm]   
             \node (fig1) at (0,0)
               {\includegraphics[width=\linewidth,trim={100px 0px 140px 0},clip]{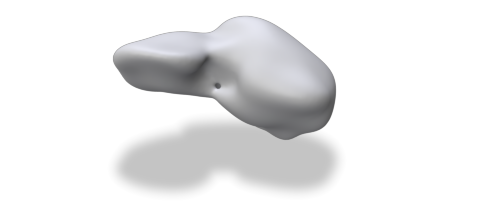}};
             \node (fig2) at (0,0.5)
               {\includegraphics[width=\insetwidth,trim={100px 50px 140px 0},clip]{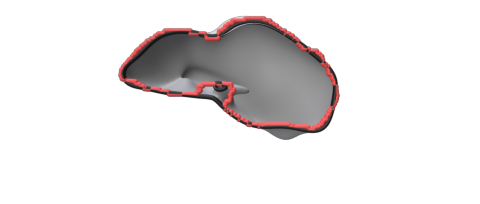}};  
        \end{tikzpicture}
        \vspace{-\snyd}
        \caption{\dmlp{} (global).}
    \end{subfigure}
    \begin{subfigure}[b]{0.245\linewidth}
        \centering
        \begin{tikzpicture}[every node/.style={anchor=south west,inner sep=0pt},x=1mm, y=1mm]   
             \node (fig1) at (0,0)
               {\includegraphics[width=\linewidth,trim={100px 0px 140px 0},clip]{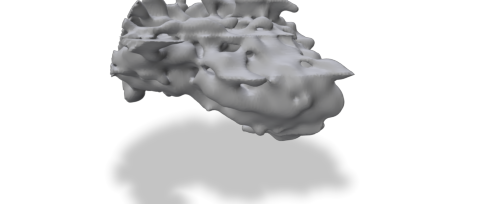}};
             \node (fig2) at (0,0.5)
               {\includegraphics[width=\insetwidth,trim={100px 50px 140px 0},clip]{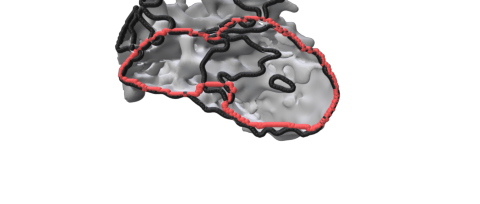}};  
        \end{tikzpicture}
        \vspace{-\snyd}
        \caption{\dmlp{} (local).}
    \end{subfigure}
    \begin{subfigure}[b]{0.245\linewidth}
        \centering
        \begin{tikzpicture}[every node/.style={anchor=south west,inner sep=0pt},x=1mm, y=1mm]   
             \node (fig1) at (0,0)
               {\includegraphics[width=\linewidth,trim={90px 0px 130px 0},clip]{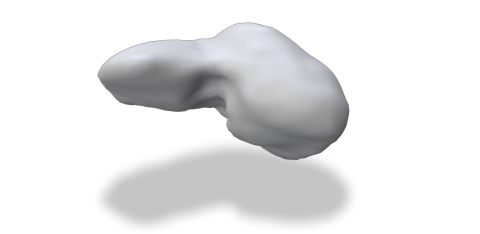}};
             \node (fig2) at (0,0.5)
               {\includegraphics[width=\insetwidth,trim={90px 50px 130px 0},clip]{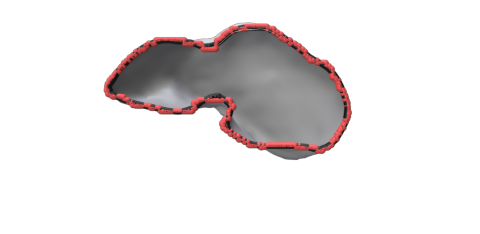}};  
        \end{tikzpicture}
        \vspace{-\snyd}
        \caption{\ours{} (Ours).}
    \end{subfigure}
    \caption{{\bf Reconstruction of a previously unseen liver from 2500 3D points.} For each method, we present the full 3D volume and a version of it cut in the middle. The red outline denotes the ground-truth section and the black one that of the reconstructed organ. Note that only ours is smooth while still following closely the ground-truth one.}
    \label{fig:liver-results}
\end{figure}
\begin{table}[ht!]
    \newcommand{\tpm}{{\tiny~}{\footnotesize$\pm$}{\tiny~}}
    \setlength{\tabcolsep}{3.1pt}
    \centering\small
    \caption{\textbf{Quantitative results for reconstructing unseen livers from unoriented points.} For each metric we report the mean and standard deviation over the reconstructed shapes. \ours{} concistently produces better reconstructions while still having very good mesh quality.}
    \label{tab:liver-recon-results}
    \begin{tabularx}{\linewidth}{Xrrrrrr}
    \toprule
    & Chamfer*$\downarrow$ & Hausdorff$\downarrow$ & F@1\%$\uparrow$ & F@2\%$\uparrow$ & Quality$\uparrow$ & \%self. ints.$\downarrow$ \\
    \midrule
    \dsdf{} \cite{park2019deepsdf} & 40.7\tpm23.7 & 0.21\tpm 0.06 & 31.3\tpm13.0 & 63.8\tpm15.6 & \textbf{0.98\tpm 0.00} & \textbf{0.00\tpm 0.00} \\
    {\it SIREN}+\dsdf{} \cite{park2019deepsdf,sitzmann2020implicit} & 36.2\tpm34.1 & 0.20\tpm0.04 & 51.4\tpm8.68 & 78.4\tpm9.27 & \textbf{0.98\tpm 0.00} & \textbf{0.00\tpm 0.00}  \\
    \dasm{} \cite{wickramasinghe2021deep} & 17.0\tpm10.0 & 0.23\tpm0.50 & 87.7\tpm3.25 & 92.9\tpm2.57 & 0.74\tpm 0.03 & 7.40\tpm4.57 \\
    \dasr{} \cite{wickramasinghe2021deep} & 8.8\tpm7.53 & 0.19\tpm0.07 & 94.6\tpm2.57 & 96.8\tpm1.97 & \textbf{0.98\tpm0.00} & \textbf{0.00\tpm0.00} \\
    \dmlp{}  (global) \cite{mehta2021modulated} & 13.6\tpm5.67 & 0.16\tpm0.05 & 71.7\tpm6.39 & 91.4\tpm3.48 & \textbf{0.98\tpm0.00} & \textbf{0.00\tpm0.00} \\
    \dmlp{} (local) \cite{mehta2021modulated} & 161.7\tpm 39.6 & 0.43\tpm 0.04 & 44.5\tpm 4.05 & 62.1\tpm 4.08 & \textbf{0.98\tpm 0.00} & \textbf{0.00\tpm 0.00} \\
    \midrule
    \ours{} (Ours) & \textbf{2.4\tpm 1.04} & \textbf{0.11\tpm 0.04} & \textbf{95.4\tpm 2.06} & \textbf{99.0\tpm 0.76} & \textbf{0.98\tpm 0.00} & 0.20\tpm 0.40 \\
    \bottomrule
    \multicolumn{7}{l}{*multiplied with 10,000}
    \end{tabularx}
    \label{tab:liverFit}
\end{table}

We report comparative results in Tab.~\ref{tab:liverFit}. \ours{} consistently outperforms the other approaches, in part because it can model sharp features more accurately, as can be seen in the qualitative results of Fig.~\ref{fig:liver-results}. Note especially the left side point and the concavity in the lower middle part of the liver. \ours{} also produces excellent mesh quality and keeps the number of intersecting triangles very low although not zero. In future work, we will add an additional loss term to eliminate them.

{
\centering

\begin{minipage}{0.41\linewidth}
    \newcommand{\tpm}{{\tiny~}{\footnotesize$\pm$}{\tiny~}}
    \centering\small
    \setlength{\tabcolsep}{3pt}
    \captionof{table}{\textbf{Quantitative results for reconstructing unseen spleens.} Metrics are reported as in Tab.~\ref{tab:liver-recon-results}.}
    \label{tab:spleen-fit-results}
    \begin{tabular}{lrr}
        \toprule
         & \dasr{} \cite{wickramasinghe2021deep} & \ours{} (Ours) \\
         \midrule
         Chamfer*$\downarrow$ & \textbf{1.6\tpm 0.05} & 2.5\tpm 0.89  \\
         Hausdorff$\downarrow$ & \textbf{0.05\tpm 0.02} & 0.06\tpm 0.01 \\ 
         F@1\%$\uparrow$ & \textbf{97.8\tpm 1.75} & 92.9\tpm 2.73 \\
         F@2\%$\uparrow$ & \textbf{100\hspace{2pt}\tpm 0.05} & 99.9\tpm 0.11 \\
         Quality$\uparrow$ & \textbf{0.98\tpm 0.00} & \textbf{0.98\tpm 0.00} \\
         \%self. ints.$\downarrow$ & \textbf{0.00\tpm 0.00} & \textbf{0.00\tpm 0.00} \\
        \bottomrule
        \multicolumn{3}{l}{*multiplied with 10,000}
    \end{tabular}
\end{minipage}
\hfill
\newcommand{\insetwidth}{0.3\linewidth}
\begin{minipage}{0.55\linewidth}
    \centering
    \begin{tabular}{cc}
    \setlength{\tabcolsep}{0pt}
        \begin{tikzpicture}[every node/.style={anchor=south west,inner sep=0pt},x=1mm, y=1mm]   
             \node (fig1) at (0,0)
               {\includegraphics[width=0.45\linewidth,trim={120px 30px 140px 0},clip]{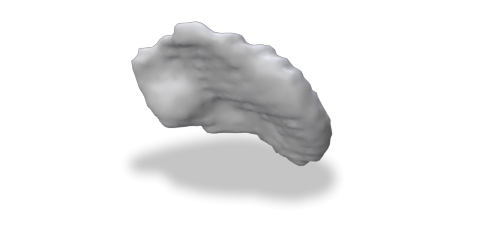}};
             \node (fig2) at (0,0)
               {\includegraphics[width=\insetwidth,trim={120px 50px 140px 0},clip]{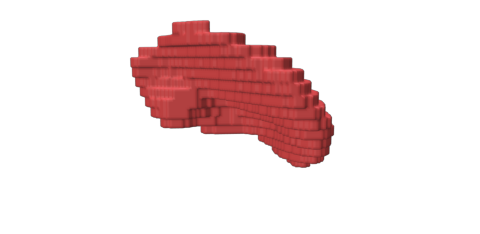}};  
        \end{tikzpicture}
        &
        \begin{tikzpicture}[every node/.style={anchor=south west,inner sep=0pt},x=1mm, y=1mm]   
             \node (fig1) at (0,0)
               {\includegraphics[width=0.45\linewidth,trim={120px 30px 140px 0},clip]{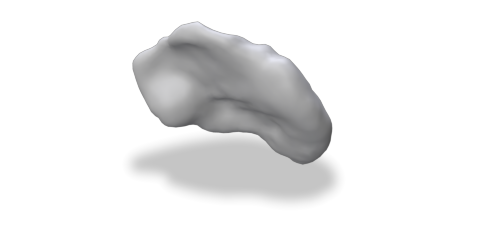}};
             \node (fig2) at (0,0)
               {\includegraphics[width=\insetwidth,trim={120px 50px 140px 0},clip]{figures/spleen_fit/spleen_noshadow_GroundTruth.png}};  
        \end{tikzpicture}\\
        (a) \dasr{} & (b) \ours{} (Ours)
    \end{tabular}
    \captionof{figure}{\textbf{Reconstruction of previously unseen spleens from 2500 3D points.} The inset shows the ground truth in red.}
    \label{fig:spleen-fit-results}
    
\end{minipage}
}


We repeated the experiment on the much simpler and smoother spleen shapes. We focused on \ours{} and \dasr{} because they delivered the best results on the liver. As can be seen in Tab.~\ref{tab:spleen-fit-results} and Fig.~\ref{fig:spleen-fit-results} \dasr{} delivers very slightly better metrics but a qualitatively worse reconstruction because it overfits to the staircase artifacts on the ground-truth shape. In contrast, \ours{} yields an organic shape which still fits the data well and can therefore be viewed as a more realistic result. This effect also exists in the liver dataset but did not affect the metrics as obviously because the original images were of higher resolution and the artifacts had much less of an effect.


\subsection{Shape Reconstruction from Planar Curve Annotations}

\paragraph{Experimental Setup.}

When annotating medical images, a common time-saving practice is to only annotate three orthogonal 2D slices instead of the entire 3D image. We test the ability of our model to reconstruct shapes from such weak annotations by fitting to the 2D planar boundary curves extracted from the 2D slice annotations. We compute the intersection curves between each held out liver and three orthogonal axis-aligned planes and sample 5,000 points randomly on each curve. As can be seen in Fig.~\ref{fig:liver-slice-results}(e), this represents a very sparse set of data points and the quality of the shape priors embedded in the models is key to obtaining good results. 

For \ours{}, \dasm{}, and \dasr{} we take $\cL_\mathrm{task}$ to be a modified Chamfer distance that relies on distances within the annotation planes (see supplementary for details). For the other methods, we again use the mean absolute SDF value to compute $\cL_\mathrm{task}$. We again use the ADAM optimizer and Botsch-Kobbelt remeshing as in the previous section. We set $\lambda_\mathrm{reg}=0.01$ and $\lambda_\mathrm{dir}=100$ as we want to rely heavily on the shape prior in this task.

\paragraph{Results.}  We report comparative results in Tab.~\ref{tab:liver-fit-planar-results} and qualitative results in Fig.~\ref{fig:liver-slice-results}. To generate these results, we only used annotations in the three orthogonal axis-aligned planes. However, we can also annotate additional planes to provide further information. In Fig.~\ref{fig:liver-slice-fit-metrics}, we plot the same quality metrics as in Tab.~\ref{tab:liver-fit-planar-results} as a function of the number of annotated planes for one of the livers. Not only are \dasm{} results consistently better, but they improve almost monotonically with the number of planes we provide, which is a very desirable behavior in clinical practice.

\begin{table}[ht!]
    \newcommand{\tpm}{{\tiny~}{\footnotesize$\pm$}{\tiny~}}
    \setlength{\tabcolsep}{3pt}
    \centering\small
    \caption{\textbf{Quantitative results for reconstructing unseen livers from planar curve annotations.} For each metric we report the mean and standard deviation over the reconstructed shapes. \ours{} consistently outperforms the baselines while retaining excellent mesh quality.}
    \label{tab:liver-fit-planar-results}
    \begin{tabularx}{\linewidth}{Xrrrrrr}
    \toprule
    & Chamfer*$\downarrow$ & Hausdorff$\downarrow$ & F@1\%$\uparrow$ & F@2\%$\uparrow$ & Quality$\uparrow$ & \%self. ints.$\downarrow$ \\
    \midrule
    \dsdf{} \cite{park2019deepsdf} & 4.36\tpm 2.35 & 0.22\tpm 0.06 & 41.5\tpm 12.1 & 69.5\tpm 12.2 & \textbf{0.98\tpm 0.00} & \textbf{0.00\tpm 0.00} \\
    \dsds{} \cite{park2019deepsdf,sitzmann2020implicit} & 3.82\tpm 2.24 & \textbf{0.21\tpm 0.05} & 47.0\tpm 7.67 & 74.4\tpm 8.25 & \textbf{0.98\tpm 0.00} & \textbf{0.00\tpm 0.00} \\
    \dasr{} \cite{wickramasinghe2021deep} & 27.41\tpm 7.91 & 0.47\tpm 0.09 & 14.8\tpm 5.99 & 29.5\tpm 9.99 & \textbf{0.98\tpm 0.00} & 0.02\tpm 0.07 \\
    \dmlp{} (global) \cite{mehta2021modulated} & 4.05\tpm 1.58 & 0.23\tpm 0.05 & 49.9\tpm 6.17 & 74.4\tpm 6.08 & \textbf{0.98\tpm 0.00} & \textbf{0.00\tpm 0.00} \\
    \dmlp{} (local) \cite{mehta2021modulated} & 15.55\tpm 4.53 & 0.39\tpm 0.05 & 28.3\tpm 2.87 & 48.3\tpm 4.20 & \textbf{0.98\tpm 0.00} & \textbf{0.00\tpm 0.00}  \\
    \midrule
    \ours{} (Ours) & \textbf{3.27\tpm 1.48} & \textbf{0.21\tpm 0.05} & \textbf{52.1\tpm 6.56} & \textbf{77.2\tpm 6.87} & \textbf{0.99\tpm 0.00} & \textbf{0.00\tpm 0.00} \\
    \bottomrule
    \multicolumn{7}{l}{*multiplied with 1,000}
    \end{tabularx}
\end{table}

\setlength{\snyd}{3\baselineskip}
\begin{figure}[h]
    \centering
    \begin{subfigure}[b]{0.245\linewidth}
        \centering
        \includegraphics[width=\linewidth,trim={90px 30px 110px 0},clip]{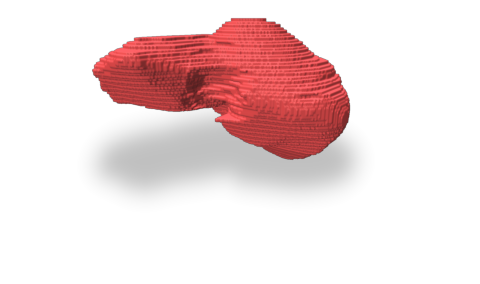}
        \vspace{-\snyd}
        \caption{Ground truth.}
    \end{subfigure}
    \begin{subfigure}[b]{0.245\linewidth}
        \centering
        \includegraphics[width=\linewidth,trim={90px 30px 110px 0},clip]{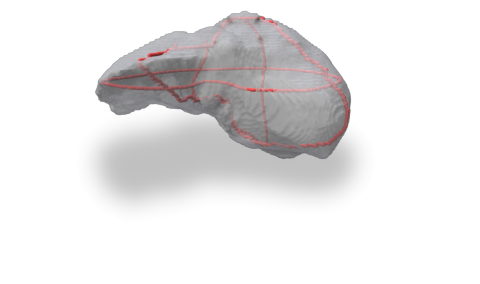}
        \vspace{-\snyd}
        \caption{\dsdf{}.}
    \end{subfigure}
    \begin{subfigure}[b]{0.245\linewidth}
        \centering
        \includegraphics[width=\linewidth,trim={90px 30px 110px 0},clip]{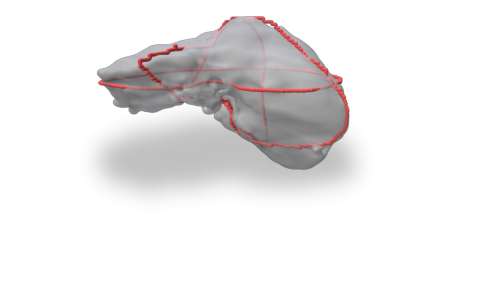}
        \vspace{-\snyd}
        \caption{\dsds{}.}
    \end{subfigure}
    \begin{subfigure}[b]{0.245\linewidth}
        \centering
        \includegraphics[width=\linewidth,trim={90px 30px 110px 0},clip]{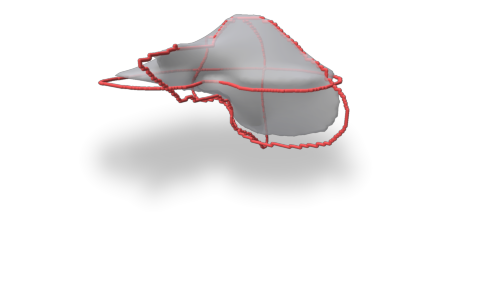}
        \vspace{-\snyd}
        \caption{\dasr{}.}
    \end{subfigure}
    \\
    \begin{subfigure}[b]{0.245\linewidth}
        \centering
        \includegraphics[width=\linewidth,trim={90px 30px 110px 0},clip]{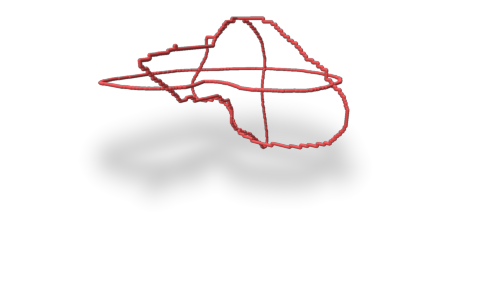}
        \vspace{-\snyd}
        \caption{Annotations.}
    \end{subfigure}
    \begin{subfigure}[b]{0.245\linewidth}
        \centering
        \includegraphics[width=\linewidth,trim={90px 30px 110px 0},clip]{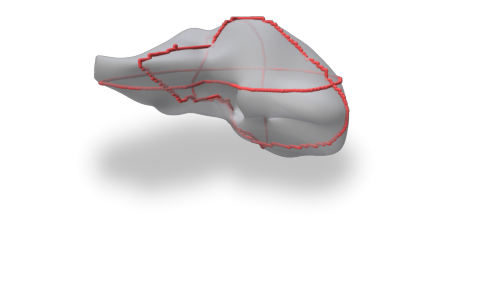}
        \vspace{-\snyd}
        \caption{\dmlp{} (global).}
    \end{subfigure}
    \begin{subfigure}[b]{0.245\linewidth}
        \centering
        \includegraphics[width=\linewidth,trim={90px 30px 110px 0},clip]{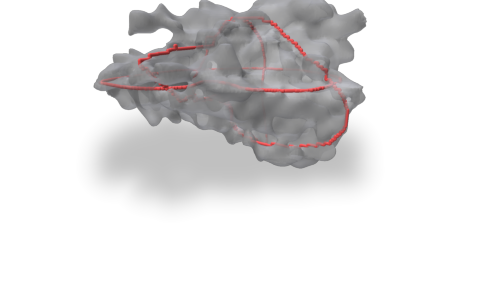}
        \vspace{-\snyd}
        \caption{\dmlp{} (local).}
    \end{subfigure}
    \begin{subfigure}[b]{0.245\linewidth}
        \centering
        \includegraphics[width=\linewidth,trim={90px 30px 110px 0},clip]{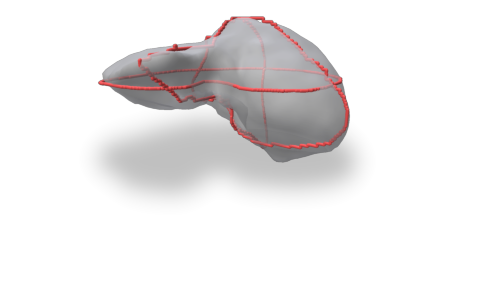}
        \vspace{-\snyd}
        \caption{\ours{} (Ours).}
    \end{subfigure}
    \caption{{\bf Reconstruction of a previously unseen liver from outlines in three different planes.} The outlines are shown in the bottom left panel. Again, our reconstruction is smooth while matching the outlines very accurately.}
    \label{fig:liver-slice-results}
\end{figure}

\begin{figure}
    \centering
    \includegraphics[width=0.24\linewidth]{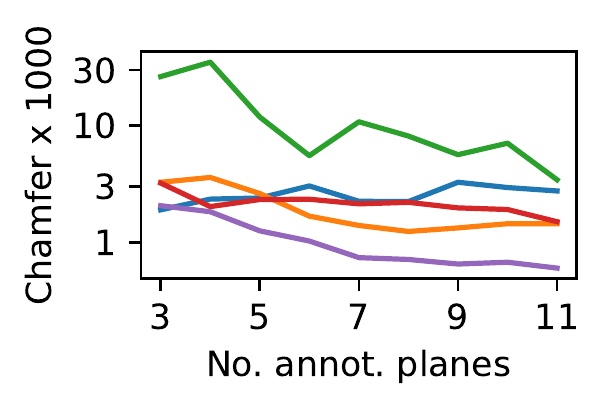}
    \includegraphics[width=0.24\linewidth]{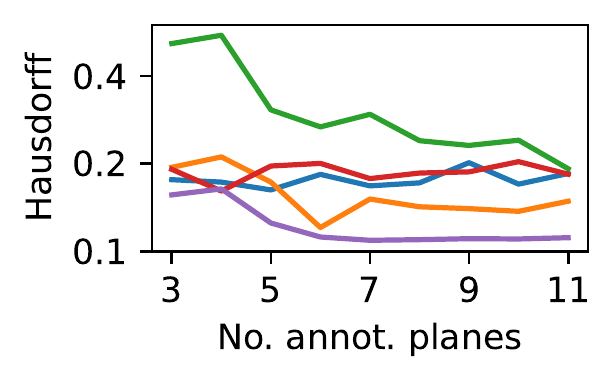}
    \includegraphics[width=0.24\linewidth]{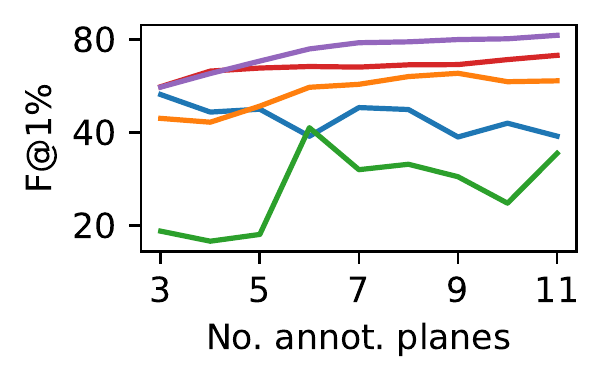}
    \includegraphics[width=0.24\linewidth]{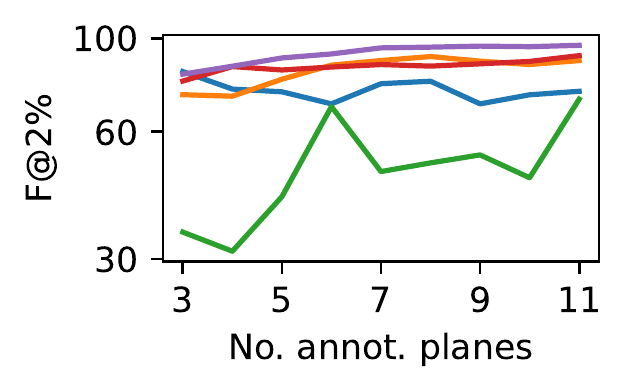}\\
    \includegraphics[width=0.9\linewidth]{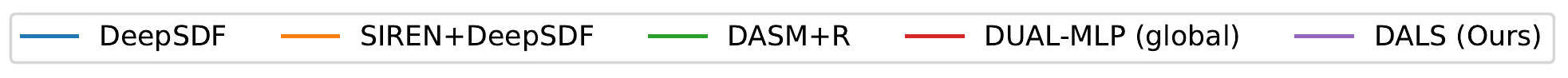}
    \caption{\textbf{Reconstruction metrics for a liver as a function of the number of annotated planes.} Unlike those of other approaches, \ours{} results, shown in purple, consistently improve as more planes are added. However, they tend to saturate after 6 or 7.}
    \label{fig:liver-slice-fit-metrics}
\end{figure}

\subsection{3D Image Segmentation with Little Training Data}

\paragraph{Experimental setup}

A common medical image analysis task is to segment objects with very few annotations available. Here, we use \ours{} to refine a voxel segmentation produced by a CNN backbone network trained on a few 2D slice annotations. As a backbone, we use the standard U-Net~\cite{cciccek20163d} and V-Net~\cite{milletari2016v}, along with the more recent nn-U-Net~\cite{isensee2021nnu} and UNETR~\cite{hatamizadeh2022unetr}.

For this experiment, we only use the 40 liver images we held off for testing in the previous experiments. We use 20 of them to train the backbone and the other 20 for testing purposes.  This simulates a realistic scenario in which we have few training images to train the segmentation network and they have {\it not} been used to learn the shape priors. 

\paragraph{Results.}

\begin{table}[ht!]
    \newcommand{\tpm}{{\tiny~}{\footnotesize$\pm$}{\tiny~}}
    \setlength{\tabcolsep}{5pt}
    \centering
    \caption{\textbf{Quantitative results for segmentation refinement.} For each metric we report the mean and standard deviation over the 20 reconstructed shapes. Refinement with \ours{} consistently improves the segmentations for all backbones and metrics.}
    \label{tab:liver-seg-results}
    \small
    \begin{tabularx}{\linewidth}{Xrrrrrr}
    \toprule
     & \multicolumn{2}{c}{Dice$\uparrow$} & \multicolumn{2}{c}{Hausdorff$\downarrow$} & \multicolumn{2}{c}{Chamfer*$\downarrow$} \\
     & \multicolumn{1}{c}{Raw} & \multicolumn{1}{c}{w/ DALS} & \multicolumn{1}{c}{Raw} & \multicolumn{1}{c}{w/ DALS} & \multicolumn{1}{c}{Raw} & \multicolumn{1}{c}{w/ DALS} \\
    \midrule
    U-Net \cite{cciccek20163d} & 0.81\tpm0.08 & \textbf{0.83\tpm0.08} & 26.6\tpm10.2 & \textbf{20.9\tpm6.32} & 44.9\tpm51.0 & \textbf{26.0\tpm22.4} \\
    V-Net \cite{milletari2016v} & 0.79\tpm0.16 & \textbf{0.80\tpm0.16} & 28.5\tpm12.5 & \textbf{25.4\tpm9.29} & 56.4\tpm74.8 & \textbf{43.3\tpm54.8} \\
    nn-U-Net \cite{isensee2021nnu} & 0.84\tpm0.09 & \textbf{0.85\tpm0.07} & 25.2\tpm11.3 & \textbf{19.5\tpm8.08} & 38.7\tpm48.4 & \textbf{22.3\tpm20.4} \\
    UNETR \cite{hatamizadeh2022unetr} & 0.74\tpm0.13 & \textbf{0.75\tpm0.17} & 39.5\tpm24.2 & \textbf{26.4\tpm12.5} & 164.5\tpm266\hspace*{2pt} & \textbf{52.7\tpm57.0} \\
    \bottomrule
    \multicolumn{7}{l}{*multiplied with 10,000}
    \end{tabularx}
\end{table}

\begin{figure}
    \centering
    \newcommand{\imwidth}{\linewidth}
    \begin{subfigure}{0.245\linewidth}
        \centering
        \includegraphics[width=\imwidth,trim={20px 10px 0px 10px},clip]{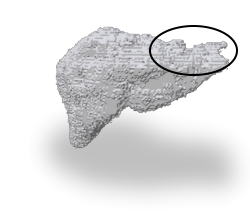}\\\vspace*{-10pt}
        \includegraphics[width=\imwidth,trim={20px 10px 0px 10px},clip]{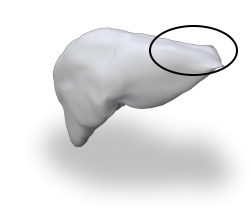}\vspace*{-10pt}
        \caption{U-Net.}
    \end{subfigure}
    \begin{subfigure}{0.245\linewidth}
        \centering
        \includegraphics[width=\imwidth,trim={20px 10px 0px 10px},clip]{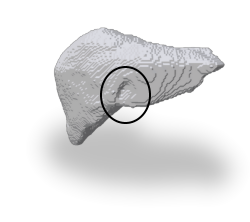}\\\vspace*{-10pt}
        \includegraphics[width=\imwidth,trim={20px 10px 0px 10px},clip]{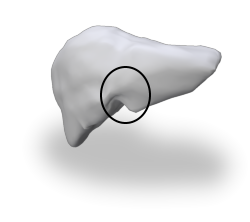}\vspace*{-10pt}
        \caption{V-Net.}
    \end{subfigure}
    \begin{subfigure}{0.245\linewidth}
        \centering
        \includegraphics[width=\imwidth,trim={20px 10px 0px 10px},clip]{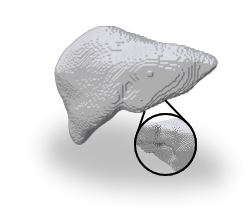}\\\vspace*{-10pt}
        \includegraphics[width=\imwidth,trim={20px 10px 0px 10px},clip]{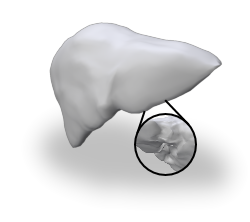}\vspace*{-10pt}
        \caption{nn-U-Net.}
    \end{subfigure}
    \begin{subfigure}{0.245\linewidth}
        \centering
        \includegraphics[width=\imwidth,trim={20px 10px 0px 10px},clip]{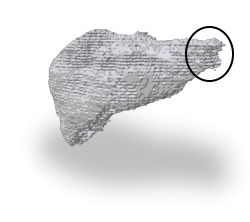}\\\vspace*{-10pt}
        \includegraphics[width=\imwidth,trim={20px 10px 0px 10px},clip]{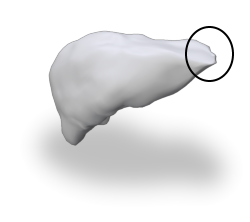}\vspace*{-10pt}
        \caption{UNETR.}
    \end{subfigure}
    \caption{\textbf{Comparison of raw (top) and refined (bottom) segmentations.} The black rings highlight examples of how refinement with \ours{} corrects segmentation mistakes.}
    \label{fig:segment-refine-results}
\end{figure}

Because the models are not sufficiently well trained --- a common occurrence in medical imaging --- the `raw' segmentations reported in Tab.~\ref{tab:liver-seg-results} are not particularly good. However, to refine them by enforcing shape priors, we can treat them as noisy data to which we fit a \ours{} model. To this end, we initialize the shape at the center and scale predicted by the raw segmentation. We then fit \ours{} to an unsigned distance function computed from the segmentation binary image (full details in supplementary). We use $\lambda_\mathrm{reg}=0$ and $\lambda_\mathrm{dir}=\infty$ to heavily rely on the model's learned prior.

As can be seen in Fig.~\ref{fig:segment-refine-results} and  Tab.~\ref{tab:liver-seg-results}, this yields much improved segmentations both in terms of visual appearance and quantitative metrics. Note that \ours{} removes spurious growths \emph{and} recovers concavities that were missed in the raw segmentations. In other words, \ours{} does not simply smooth. It really enforces geometric priors. 

\subsection{Ablation experiments}

\begin{figure}
    \centering
    \setlength{\snyd}{1.8\baselineskip}
    \newcommand{\insetwidth}{0.5\linewidth}
    \begin{subfigure}[b]{0.23\linewidth}
        \centering
        \begin{tikzpicture}[every node/.style={anchor=south west,inner sep=0pt},x=1mm, y=1mm]   
             \node (fig1) at (0,0)
               {\includegraphics[width=\linewidth,trim={90px 0px 100px 0},clip]{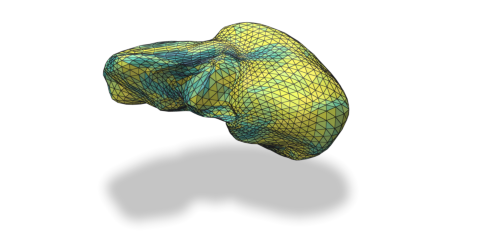}};
             \node (fig2) at (0,0.5)
               {\includegraphics[width=\insetwidth,trim={100px 50px 140px 0},clip]{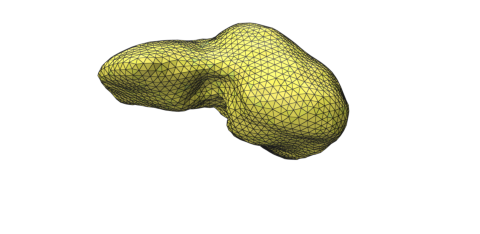}};  
        \end{tikzpicture}
        \vspace{-\snyd}
        \caption{Many latent vectors \\ and $\lambda_\mathrm{reg}= 0$.}
    \end{subfigure}
    \begin{subfigure}[b]{0.23\linewidth}
        \centering
        \begin{tikzpicture}[every node/.style={anchor=south west,inner sep=0pt},x=1mm, y=1mm]   
             \node (fig1) at (0,0)
               {\includegraphics[width=\linewidth,trim={90px 0px 100px 0},clip]{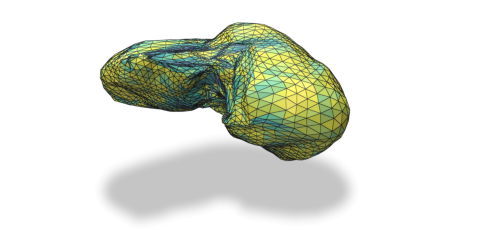}};
             \node (fig2) at (0,0.5)
               {\includegraphics[width=\insetwidth,trim={100px 50px 140px 0},clip]{figures/ablations/ablation_qual_noshadow_Meshdecoder_full.png}};  
        \end{tikzpicture}
        \vspace{-\snyd}
        \caption{Single latent vector \\ and $\lambda_\mathrm{reg}= 0$}
    \end{subfigure}
    \begin{subfigure}[b]{0.23\linewidth}
        \centering
        \begin{tikzpicture}[every node/.style={anchor=south west,inner sep=0pt},x=1mm, y=1mm]   
             \node (fig1) at (0,0)
               {\includegraphics[width=\linewidth,trim={90px 0px 100px 0},clip]{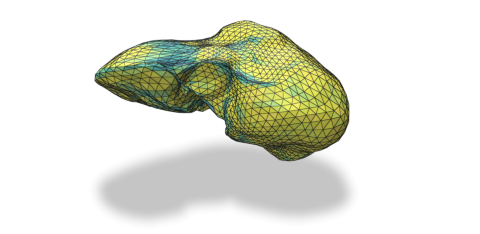}};
             \node (fig2) at (0,0.5)
               {\includegraphics[width=\insetwidth,trim={100px 50px 140px 0},clip]{figures/ablations/ablation_qual_noshadow_Meshdecoder_full.png}};  
        \end{tikzpicture}
        \vspace{-\snyd}
        \caption{Many latent vector \\ and $\lambda_\mathrm{reg}\neq 0$.}
    \end{subfigure}
    \begin{subfigure}[b]{0.23\linewidth}
        \centering
        \begin{tikzpicture}[every node/.style={anchor=south west,inner sep=0pt},x=1mm, y=1mm]   
             \node (fig1) at (0,0)
               {\includegraphics[width=\linewidth,trim={90px 0px 100px 0},clip]{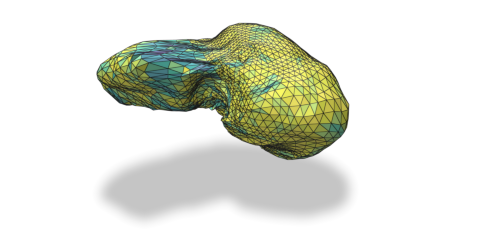}};
             \node (fig2) at (0,0.5)
               {\includegraphics[width=\insetwidth,trim={100px 50px 140px 0},clip]{figures/ablations/ablation_qual_noshadow_Meshdecoder_full.png}};  
        \end{tikzpicture}
        \vspace{-\snyd}
        \caption{Single latent vector \\ and $\lambda_\mathrm{reg}\neq 0$.}
    \end{subfigure}%
    \begin{tikzpicture}
         \node at (0,0) {\includegraphics[height=0.26\linewidth]{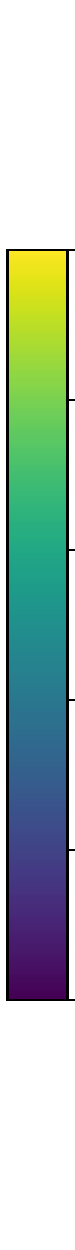}};
         \node at (0.3,-1.1){\tiny 0.0};
         \node at (0.3, 1.1){\tiny 1.0};
         \node at (0.3, -0.66){\tiny 0.2};
         \node at (0.3, -0.22){\tiny 0.4};
         \node at (0.3, 0.22){\tiny 0.6};
         \node at (0.3, 0.66){\tiny 0.8};
    \end{tikzpicture}
    
    \caption{\textbf{Reconstructions of the ablated models of \cref{tab:ablation-results}}. Facets are colored according to their quality in terms of the measure of Eq.~\eqref{eq:smooth}: yellow is high and blue is low. The insets show the reconstruction by \ours{} including re-meshing.}
    \label{fig:ablations}
\end{figure}

We perform an ablation study to investigate how using a single or multiple latent vectors and our triangle quality loss $L_\mathrm{reg}$ affect performance. To use a single vector, we constrain all latent vectors to be the same during fitting. To remove $L_\mathrm{reg}$ we set $\lambda_\mathrm{reg}$ to 0 during training and fitting.



\begin{table}[h]
    \newcommand{\tpm}{{\tiny~}{\footnotesize$\pm$}{\tiny~}}
    \setlength{\tabcolsep}{3pt}
    \centering
    \small
    \caption{\textbf{Quantitative results of the ablation study.} Local inference results in a large boost to reconstruction accuracy and the $L_\mathrm{reg}$ loss significantly improves triangle quality. Adding remeshing further boosts the triangle quality at some expense to reconstruction accuracy.}
    \label{tab:ablation-results}
    \begin{tabular}{ccrr}
        \toprule
        Local & $L_\mathrm{reg}$ & Chamfer*$\downarrow$ & Quality$\uparrow$ \\
        \midrule
         & & 7.89\tpm 2.75 & 0.75\tpm 0.03  \\
         \checkmark & & 1.89\tpm 0.63 & 0.72\tpm 0.02 \\
         & \checkmark & 5.41\tpm 1.77 & 0.83\tpm 0.02 \\
        \bottomrule
    \end{tabular}
    \qquad
    \begin{tabular}{cccrr}
        \toprule
        Local & $L_\mathrm{reg}$ & Remeshing & Chamfer*$\downarrow$ & Quality$\uparrow$ \\
        \midrule
         \checkmark & \checkmark & & 1.50\tpm 0.49 & 0.87\tpm 0.01 \\
         \checkmark & \checkmark & \checkmark & 2.41\tpm 1.04 & 0.98\tpm 0.00 \\
        \bottomrule
    \end{tabular}
\end{table}

As shown in \cref{tab:ablation-results}, both our local latent vector approach and $L_\mathrm{reg}$ loss significantly improves reconstruction and mesh quality, even more so when combined. This is also apparent in \cref{fig:ablations} as the reconstructions with $L_\mathrm{reg}$ are smoother and fewer triangles are severely distorted. Finally, adding remeshing results in excellent mesh quality at some cost to reconstruction accuracy. We chose to prioritise the mesh quality for our results in this work.


\section{Conclusion}

We have shown that we train a latent vector model to represent a complex surface by a triangulated sphere and deformations at each one of its vertices that are the output of a decoder that takes as input the vertex coordinates and a latent vector. At training time, we use an auto-decoder approach to learn a single latent vector per training shape. However, at model-fitting time, we allow the latent vectors to be different at each vertex but we enforce consistency of these vectors across the triangulation. This enables us to learn the model from a relatively small training set while giving the necessary flexibility to model complex 3D shapes without over-smoothing. A key ingredient is that we impose regularization constraints on the latent vectors but {\it not} on the vertex 3D locations. 

In this work, we have focused on organic shapes represented as watertight surfaces and demonstrated the effectiveness of our approach on liver and spleen reconstruction. In future work, we will extend our approach to non-watertight surfaces by dynamically updating the template mesh, that is, by removing faces that are predicted to be unused as in~\cite{pan2019deep}. We will also integrate re-meshing into \ours{} as we did for \dasr{} to further improve mesh quality. 

Finally, as the purpose of our model is to incorporate prior knowledge into medical image processing tasks, it is important to mention that our model may introduce unwanted biases if the training data is heavily skewed towards certain genders, ages, or other groups. The local nature of our model makes this bias less direct, but representative training data is still a must.

\printbibliography

@String { AI           = {{Artificial Intelligence}} }

@String { CVPR         = {{Proceedings of the {IEEE} Conference on Computer Vision and Pattern Recognition ({CVPR})}} }

@String { ECCV         = {{Proceedings of the European Conference on Computer Vision ({ECCV})}} }

@String { EG           = {{Proceedings of Eurographics}} }

@String { ICCV         = {{Proceedings of the International Conference on Computer Vision ({ICCV})}} }

@String { ICLR         = {{Proceedings of the International Conference on Learning Representations ({ICLR})}} }

@String { ICML         = {{Proceedings of the International Conference on Machine Learning ({ICML})}} }

@String { IJCV         = {{International Journal of Computer Vision ({IJCV})}} }

@String { IVC          = {{Image and Vision Computing ({IVC})}} }

@String { MICCAI       = {{Proceedings of the International Conference on Medical Image Computing and Computer Assisted Intervention ({MICCAI})}} }

@String { NETWORKS     = {{Networks}} }

@String { NIPS         = {{Neural Information Processing Systems ({NeurIPS})}} }

@String { OR           = {{Operations Research}} }

@String { THREEDV      = {{Proceedings of the International Conference on 3D Vision ({3DV})}} }

@String { TOG          = {{{{ACM} Transactions on Graphics ({TOG})}}} }

@String { WCACV        = {{Proceedings of the IEEE Winter Conference on Applications of Computer Vision ({WACV})}} }

@inproceedings{pan2019deep,
  title={Deep mesh reconstruction from single rgb images via topology modification networks},
  author={Pan, Junyi and Han, Xiaoguang and Chen, Weikai and Tang, Jiapeng and Jia, Kui},
  booktitle=CVPR,
  pages={9964--9973},
  year={2019}
}

@inproceedings{botsch2004remeshing,
  title={A remeshing approach to multiresolution modeling},
  author={Botsch, Mario and Kobbelt, Leif},
  booktitle=EG,
  pages={185--192},
  year={2004}
}

@article{fua1995object,
  author = {P. Fua and Y. G. Leclerc},
  title = {{Object-Centered Surface Reconstruction: Combining Multi-Image Stereo and Shading}},
  journal = IJCV,
  volume = "16",
  pages = "35--56",
  month = "September",
  year = 1995
}

@inproceedings{fua1996model,
  author = {P. Fua},
  title = {{Model-Based Optimization: Accurate and Consistent Site Modeling}},
  booktitle = {Congress of International Society for Photogrammetry and Remote Sensing},
  month = "July",
  year = 1996
}

@inproceedings{hatamizadeh2020endtoend,
  author = {A. Hatamizadeh and D. Sengupta and D. Terzopoulos},
  title = {{End-To-End Trainable Deep Active Contour Models for Automated Image Segmentation: Delineating Buildings in Aerial Imagery}},
  booktitle = ECCV,
  year = 2020
}

@inproceedings{leventon2000stastical,
  author = {M. E. Leventon and W. E. Grimson and O. Faugeras},
  title = {{Statistical Shape Influence in Geodesic Active Contours}},
  booktitle = CVPR,
  pages = "316--323",
  year = 2000
}

@article{mcinerney1995dynamic,
  author = {T. Mcinerney and D. Terzopoulos},
  title = {{A Dynamic Finite Element Surface Model for Segmentation and Tracking in Multidimensional Medical Images with Application to Cardiac 4D Image Analysis}},
  journal = {Computerized Medical Imaging and Graphics},
  volume = "19",
  number = "1",
  pages = "69--83",
  year = 1995
}

@article{prevost2013incorporating,
  author = {R. Prevost and R. Cuingnet and B. Mory and B. Cohen and R. Ardon},
  title = {{Incorporating Shape Variability in Image Segmentation via Implicit Template Deformation}},
  journal = MICCAI,
  pages = "82--89",
  year = 2013
}

@inproceedings{tatarchenko2019single,
  title={What do single-view 3d reconstruction networks learn?},
  author={Tatarchenko, Maxim and Richter, Stephan R and Ranftl, Ren{\'e} and Li, Zhuwen and Koltun, Vladlen and Brox, Thomas},
  booktitle=CVPR,
  pages={3405--3414},
  year={2019}
}

@inproceedings{solomon2014laplace,
  title={Laplace-Beltrami: The Swiss army knife of geometry processing},
  author={Solomon, Justin and Crane, Keegan and Vouga, Etienne},
  booktitle={Symposium on Geometry Processing Graduate School (Cardiff, UK, 2014)},
  volume={2},
  year={2014}
}

@article{knapitsch2017tanks,
  title={Tanks and temples: Benchmarking large-scale scene reconstruction},
  author={Knapitsch, Arno and Park, Jaesik and Zhou, Qian-Yi and Koltun, Vladlen},
  journal=TOG,
  volume={36},
  number={4},
  pages={1--13},
  year={2017}
}

@article{lewiner2003efficient,
  title={Efficient implementation of marching cubes' cases with topological guarantees},
  author={Lewiner, Thomas and Lopes, H{\'e}lio and Vieira, Ant{\^o}nio Wilson and Tavares, Geovan},
  journal=JGT,
  volume={8},
  pages={1--15},
  year={2003}
}

@article{shewchuk2002good,
  title={What is a good linear finite element? interpolation, conditioning, anisotropy, and quality measures (preprint)},
  author={Shewchuk, Jonathan},
  journal={University of California at Berkeley},
  year={2002}
}

@inproceedings{smith2019geometrics,
  title     = {GEOMetrics: Exploiting Geometric Structure for Graph-Encoded Objects},
  author    = {Edward J. Smith and Scott Fujimoto and Adriana Romero and David Meger},
  booktitle = ICML,
  pages     = {5866--5876},
  year      = {2019}
}

@article{ba2016layer,
  title={Layer normalization},
  author={Ba, Jimmy Lei and Kiros, Jamie Ryan and Hinton, Geoffrey E},
  journal={arXiv preprint arXiv:1607.06450},
  year={2016}
}

@inproceedings{chabra2020deep,
  title={Deep local shapes: Learning local sdf priors for detailed 3d reconstruction},
  author={Chabra, Rohan and Lenssen, Jan E and Ilg, Eddy and Schmidt, Tanner and Straub, Julian and Lovegrove, Steven and Newcombe, Richard},
  booktitle=ECCV,
  pages={608--625},
  year={2020}
}

@inproceedings{park2019deepsdf,
  title={Deepsdf: Learning continuous signed distance functions for shape representation},
  author={Park, Jeong Joon and Florence, Peter and Straub, Julian and Newcombe, Richard and Lovegrove, Steven},
  booktitle=CVPR,
  pages={165--174},
  year={2019}
}

@inproceedings{peng2021shape,
  author={Peng, Songyou and Jiang, Chiyu "Max" and Liao, Yiyi and Niemeyer, Michael and Pollefeys, Marc and Geiger, Andreas},
  title={Shape As Points: A Differentiable Poisson Solver},
  booktitle=NIPS,
  year={2021}
}

@article{muller2022instant,
  title={Instant Neural Graphics Primitives with a Multiresolution Hash Encoding},
  author={M{\"u}ller, Thomas and Evans, Alex and Schied, Christoph and Keller, Alexander},
  journal={arXiv preprint arXiv:2201.05989},
  year={2022}
}

@inproceedings{cciccek20163d,
  title={3D U-Net: learning dense volumetric segmentation from sparse annotation},
  author={{\c{C}}i{\c{c}}ek, {\"O}zg{\"u}n and Abdulkadir, Ahmed and Lienkamp, Soeren S and Brox, Thomas and Ronneberger, Olaf},
  booktitle=MICCAI,
  pages={424--432},
  year={2016},
  organization={Springer}
}

@inproceedings{milletari2016v,
  title={V-net: Fully convolutional neural networks for volumetric medical image segmentation},
  author={Milletari, Fausto and Navab, Nassir and Ahmadi, Seyed-Ahmad},
  booktitle=THREEDV,
  pages={565--571},
  year={2016},
  organization={IEEE}
}

@inproceedings{mehta2021modulated,
  title={Modulated periodic activations for generalizable local functional representations},
  author={Mehta, Ishit and Gharbi, Micha{\"e}l and Barnes, Connelly and Shechtman, Eli and Ramamoorthi, Ravi and Chandraker, Manmohan},
  booktitle=CVPR,
  pages={14214--14223},
  year={2021}
}

@inproceedings{sitzmann2020implicit,
  title={Implicit neural representations with periodic activation functions},
  author={Sitzmann, Vincent and Martel, Julien and Bergman, Alexander and Lindell, David and Wetzstein, Gordon},
  booktitle=NIPS,
  volume={33},
  pages={7462--7473},
  year={2020}
}

@inproceedings{wickramasinghe2021deep,
  title={Deep active surface models},
  author={Wickramasinghe, Udaranga and Fua, Pascal and Knott, Graham},
  booktitle=CVPR,
  pages={11652--11661},
  year={2021}
}

@article{kass1988snakes,
  title={Snakes: Active contour models},
  author={Kass, Michael and Witkin, Andrew and Terzopoulos, Demetri},
  journal=IJCV,
  volume={1},
  number={4},
  pages={321--331},
  year={1988}
}

@inproceedings{groueix2018papier,
  title={A papier-m{\^a}ch{\'e} approach to learning 3d surface generation},
  author={Groueix, Thibault and Fisher, Matthew and Kim, Vladimir G and Russell, Bryan C and Aubry, Mathieu},
  booktitle=CVPR,
  pages={216--224},
  year={2018}
}

@article{hanocka2020point2mesh,
  title={Point2Mesh: a self-prior for deformable meshes},
  author={Hanocka, Rana and Metzer, Gal and Giryes, Raja and Cohen-Or, Daniel},
  journal=TOG,
  volume={39},
  number={4},
  pages={126--1},
  year={2020}
}

@article{isensee2021nnu,
  title={nnU-Net: a self-configuring method for deep learning-based biomedical image segmentation},
  author={Isensee, Fabian and Jaeger, Paul F and Kohl, Simon AA and Petersen, Jens and Maier-Hein, Klaus H},
  journal={Nature methods},
  volume={18},
  number={2},
  pages={203--211},
  year={2021}
}

@article{antonelli2021medical,
  title={The medical segmentation decathlon},
  author={Antonelli, Michela and Reinke, Annika and Bakas, Spyridon and Farahani, Keyvan and Landman, Bennett A and Litjens, Geert and Menze, Bjoern and Ronneberger, Olaf and Summers, Ronald M and van Ginneken, Bram and others},
  journal={arXiv preprint arXiv:2106.05735},
  year={2021}
}

@misc{williams2021nkf,
  title={Neural Fields as Learnable Kernels for 3D Reconstruction}, 
      author={Francis Williams and Zan Gojcic and Sameh Khamis and Denis Zorin and Joan Bruna and Sanja Fidler and Or Litany},
  year={2021},
  eprint={2111.13674},
  archivePrefix={arXiv},
  primaryClass={cs.CV}
}

@article{gao2020learning,
  title={Learning deformable tetrahedral meshes for 3d reconstruction},
  author={Gao, Jun and Chen, Wenzheng and Xiang, Tommy and Jacobson, Alec and McGuire, Morgan and Fidler, Sanja},
  journal=NIPS,
  volume={33},
  pages={9936--9947},
  year={2020}
}

@inproceedings{shen2021dmtet,
  title = {Deep Marching Tetrahedra: a Hybrid Representation for High-Resolution 3D Shape Synthesis},
  author = {Tianchang Shen and Jun Gao and Kangxue Yin and Ming-Yu Liu and Sanja Fidler},
  year = {2021},
  booktitle = NIPS
}

@inproceedings{peng2020deep,
  title={Deep snake for real-time instance segmentation},
  author={Peng, Sida and Jiang, Wen and Pi, Huaijin and Li, Xiuli and Bao, Hujun and Zhou, Xiaowei},
  booktitle=CVPR,
  pages={8533--8542},
  year={2020}
}

@article{terzopoulos1988constraints,
  title={Constraints on deformable models: Recovering 3D shape and nonrigid motion},
  author={Terzopoulos, Demetri and Witkin, Andrew and Kass, Michael},
  journal=AI,
  volume={36},
  number={1},
  pages={91--123},
  year={1988},
  publisher={Elsevier}
}

@article{terzopoulos1988symmetry,
  title={Symmetry-seeking models and 3D object reconstruction},
  author={Terzopoulos, Demetri and Witkin, Andrew and Kass, Michael},
  journal=IJCV,
  volume={1},
  number={3},
  pages={211--221},
  year={1988},
  publisher={Springer}
}

@inproceedings{marcos2018learning,
  title={Learning deep structured active contours end-to-end},
  author={Marcos, Diego and Tuia, Devis and Kellenberger, Benjamin and Zhang, Lisa and Bai, Min and Liao, Renjie and Urtasun, Raquel},
  booktitle=CVPR,
  pages={8877--8885},
  year={2018}
}

@inproceedings{ling2019fast,
  title={Fast interactive object annotation with curve-gcn},
  author={Ling, Huan and Gao, Jun and Kar, Amlan and Chen, Wenzheng and Fidler, Sanja},
  booktitle=CVPR,
  pages={5257--5266},
  year={2019}
}

@inproceedings{liang2020polytransform,
  title={Polytransform: Deep polygon transformer for instance segmentation},
  author={Liang, Justin and Homayounfar, Namdar and Ma, Wei-Chiu and Xiong, Yuwen and Hu, Rui and Urtasun, Raquel},
  booktitle=CVPR,
  pages={9131--9140},
  year={2020}
}

@article{nicolet2021large,
  title={Large steps in inverse rendering of geometry},
  author={Nicolet, Baptiste and Jacobson, Alec and Jakob, Wenzel},
  journal=TOG,
  volume={40},
  number={6},
  pages={1--13},
  year={2021}
}

@inproceedings{litany2018deformable,
  title={Deformable shape completion with graph convolutional autoencoders},
  author={Litany, Or and Bronstein, Alex and Bronstein, Michael and Makadia, Ameesh},
  booktitle=CVPR,
  pages={1886--1895},
  year={2018}
}

@article{morreale2022neural,
  title={Neural Convolutional Surfaces},
  author={Morreale, Luca and Aigerman, Noam and Guerrero, Paul and Kim, Vladimir G and Mitra, Niloy J},
  journal={arXiv preprint arXiv:2204.02289},
  year={2022}
}

@inproceedings{kim2021point,
  title={Point Cloud Augmentation with Weighted Local Transformations},
  author={Kim, Sihyeon and Lee, Sanghyeok and Hwang, Dasol and Lee, Jaewon and Hwang, Seong Jae and Kim, Hyunwoo J},
  booktitle=ICCV,
  pages={548--557},
  year={2021}
}

@article{bhatia1990two,
  title={Two-dimensional finite element mesh generation based on stripwise automatic triangulation},
  author={Bhatia, R. P. and Lawrence, Kent L.},
  journal={Computers \& Structures},
  volume={36},
  number={2},
  pages={309--319},
  year={1990}
}

@inproceedings{mescheder2019occupancy,
  title={Occupancy networks: Learning 3d reconstruction in function space},
  author={Mescheder, Lars and Oechsle, Michael and Niemeyer, Michael and Nowozin, Sebastian and Geiger, Andreas},
  booktitle=CVPR,
  pages={4460--4470},
  year={2019}
}

@article{takikawa2021nglod,
    title = {Neural Geometric Level of Detail: Real-time Rendering with Implicit {3D} Shapes},
    author = {Towaki Takikawa and
              Joey Litalien and 
              Kangxue Yin and 
              Karsten Kreis and 
              Charles Loop and 
              Derek Nowrouzezahrai and 
              Alec Jacobson and 
              Morgan McGuire and 
              Sanja Fidler},
    booktitle = CVPR,
    year = {2021}
}

@inproceedings{chibane2020implicit,
  title={Implicit functions in feature space for 3d shape reconstruction and completion},
  author={Chibane, Julian and Alldieck, Thiemo and Pons-Moll, Gerard},
  booktitle=CVPR,
  pages={6970--6981},
  year={2020}
}

@inproceedings{peng2020convolutional,
  title={Convolutional occupancy networks},
  author={Peng, Songyou and Niemeyer, Michael and Mescheder, Lars and Pollefeys, Marc and Geiger, Andreas},
  booktitle=ECCV,
  pages={523--540},
  year={2020}
}

@inproceedings{jiang2020local,
  title={Local implicit grid representations for 3d scenes},
  author={Jiang, Chiyu and Sud, Avneesh and Makadia, Ameesh and Huang, Jingwei and Nie{\ss}ner, Matthias and Funkhouser, Thomas and others},
  booktitle=CVPR,
  pages={6001--6010},
  year={2020}
}

@inproceedings{chen2019learning,
  title={Learning implicit fields for generative shape modeling},
  author={Chen, Zhiqin and Zhang, Hao},
  booktitle=CVPR,
  pages={5939--5948},
  year={2019}
}

@inproceedings{liu2020neural,
  title={Neural sparse voxel fields},
  author={Liu, Lingjie and Gu, Jiatao and Zaw Lin, Kyaw and Chua, Tat-Seng and Theobalt, Christian},
  booktitle=NIPS,
  pages={15651--15663},
  year={2020}
}

@article{martel2021acorn,
author = {Martel, Julien N.P. and Lindell, David B. and Lin, Connor Z. and Chan, Eric R. and Monteiro, Marco and Wetzstein, Gordon},
title = {ACORN: Adaptive Coordinate Networks for Neural Representation},
journal = TOG,
year={2021}
}

@inproceedings{achlioptas2018learning,
  title={Learning representations and generative models for 3d point clouds},
  author={Achlioptas, Panos and Diamanti, Olga and Mitliagkas, Ioannis and Guibas, Leonidas},
  booktitle=ICML,
  pages={40--49},
  year={2018}
}

@article{brock2016generative,
  title={Generative and discriminative voxel modeling with convolutional neural networks},
  author={Brock, Andrew and Lim, Theodore and Ritchie, James M and Weston, Nick},
  journal={arXiv preprint arXiv:1608.04236},
  year={2016}
}

@inproceedings{dai2017shape,
  title={Shape completion using 3d-encoder-predictor cnns and shape synthesis},
  author={Dai, Angela and Ruizhongtai Qi, Charles and Nie{\ss}ner, Matthias},
  booktitle=CVPR,
  pages={5868--5877},
  year={2017}
}

@inproceedings{kingma2014adam,
  title={Adam: A method for stochastic optimization},
  author={Kingma, Diederik P and Ba, Jimmy},
  booktitle=ICLR,
  year={2015}
}

@article{he2008comparative,
  title={A comparative study of deformable contour methods on medical image segmentation},
  author={He, Lei and Peng, Zhigang and Everding, Bryan and Wang, Xun and Han, Chia Y and Weiss, Kenneth L and Wee, William G},
  journal=IVC,
  volume={26},
  number={2},
  pages={141--163},
  year={2008}
}

@article{jorstad2015neuromorph,
  title={NeuroMorph: a toolset for the morphometric analysis and visualization of 3D models derived from electron microscopy image stacks},
  author={Jorstad, Anne and Nigro, Biagio and Cali, Corrado and Wawrzyniak, Marta and Fua, Pascal and Knott, Graham},
  journal={Neuroinformatics},
  volume={13},
  number={1},
  pages={83--92},
  year={2015}
}

@inproceedings{hatamizadeh2022unetr,
  title={UNETR: Transformers for 3d medical image segmentation},
  author={Hatamizadeh, Ali and Tang, Yucheng and Nath, Vishwesh and Yang, Dong and Myronenko, Andriy and Landman, Bennett and Roth, Holger R and Xu, Daguang},
  booktitle=WCACV,
  pages={574--584},
  year={2022}
}

@InProceedings{juhl2021implicit,
author="Juhl, Kristine Aavild and Morales, Xabier and de Backer, Ole and Camara, Oscar and Paulsen, Rasmus Reinhold",
title="Implicit Neural Distance Representation for Unsupervised and Supervised Classification of Complex Anatomies",
booktitle=MICCAI,
year="2021",
pages="405--415",
}
\vfill

\pagebreak
\appendix

We provide additional details on how we fit our models to planar curves and to voxel segmentations, as described in Sections 4.2 and 4.3 respectively. 

\section{Fitting \ours{} to Planar Curve Annotations}
When annotations are only provided in 2D planes, we only wish to evaluate the reconstruction in these planes. This is similar to how plane annotations are handled for 3D voxel segmentations \cite{cciccek20163d}.

\begin{wrapfigure}{R}{0.5\linewidth}
    \centering
    \includegraphics[width=\linewidth]{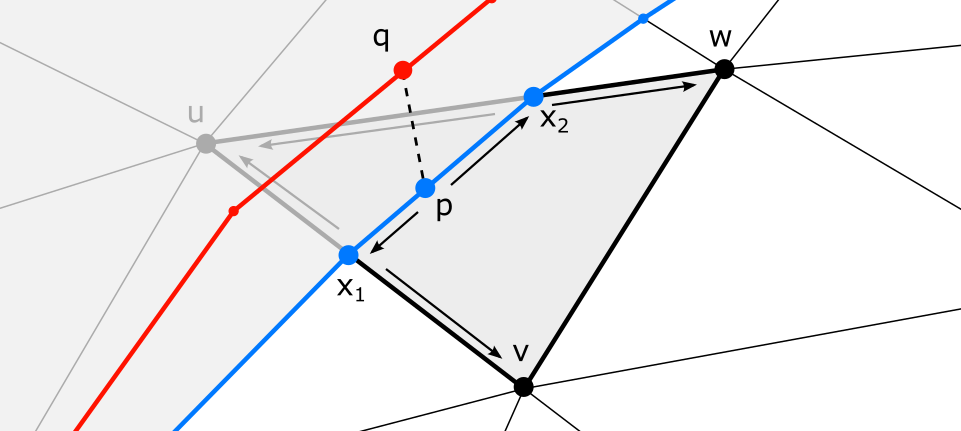} 
    \caption{\textbf{Gradient propagation for samples on the intersection between a triangle mesh and a plane.} The blue curve is the intersection of the mesh and the plane, and the red curve is the ground truth boundary curve. The highlighted triangle has vertices $\textbf{u}$, $\textbf{v}$, and $\textbf{w}$ and intersects the plane in the line segment spanned by $\textbf{x}_1$ and $\textbf{x}_2$. 
    }
    \label{fig:slice-loss}
\end{wrapfigure}
Formally, assume we are given a plane $\mathcal{P}$ and a triangle mesh $\mathcal{M}$. To differentially sample points on the intersection between $\mathcal{P}$ and $\mathcal{M}$ we first find the intersection between the plane and each triangle facet. The intersection of a plane and triangle is either empty or a line segment spanned by two points $\textbf{x}_1$ and $\textbf{x}_2$. We ignore the degenerate cases where the intersection is the entire triangle or only one of its vertices. 
We can then sample a point $\textbf{p}$ on the intersecting line segment as $\textbf{p} = r \textbf{x}_1 + (1 - r) \textbf{x}_2$, where $r \in U(0, 1)$. Let $(\alpha_1, \beta_1, \gamma_1)$ and $(\alpha_2,\beta_2,\gamma_2)$ be the barycentric coordinates of, respectively, $\textbf{x}_1$ and $\textbf{x}_2$.  We can then write $\textbf{p}$ in terms of the triangle vertices $\textbf{u}$, $\textbf{v}$, and $\textbf{w}$ as
\begin{equation}
    \textbf{p} = \begin{bmatrix}
        \textbf{u} & \textbf{v} & \textbf{w}
    \end{bmatrix}\begin{bmatrix}
        \alpha_1 & \alpha_2 \\
        \beta_1 & \beta_2 \\
        \gamma_1 & \gamma_2 
    \end{bmatrix} \begin{bmatrix}
        r \\ 1 - r
    \end{bmatrix}\,.
\end{equation}
As a result, $\textbf{p}$ is a linear combination of the triangle vertices and $r$ is an independent stochastic term. Therefore, we can propagate a gradient from the point $\textbf{p}$ to the triangle vertices $\textbf{u}$, $\textbf{v}$, and $\textbf{w}$ \cite{smith2019geometrics}, see \cref{fig:slice-loss}. 

Now, let $\mathcal{S}_\mathcal{P}(\mathcal{M})$ denote a set of $M$ points sampled differentially on the intersection of $\mathcal{M}$ and $\mathcal{P}$. Further, let $\mathcal{T}_\mathcal{P}$ be a set of $N$ points sampled uniformly on the planar curve annotations for plane $\mathcal{P}$. In this work we use $M = 5000$ sample points.
The loss for plane $\mathcal{P}$ is then
\begin{equation}
    L_\text{cf}(\mathcal{S}_\mathcal{P}(\mathcal{M}), \mathcal{T}_\mathcal{P}) =
    \frac{1}{M}\sum_{\textbf{q} \in \mathcal{T}_\mathcal{P}} \min_{\textbf{p} \in \mathcal{S}_\text{P}(\mathcal{M})} \norm{\textbf{p} - \textbf{q}}_2^2 +
    \frac{1}{M}\sum_{\textbf{p} \in \mathcal{S}_\text{P}(\mathcal{M})} \min_{\textbf{q} \in \mathcal{T}_\mathcal{P}} \norm{\textbf{q} - \textbf{p}}_2^2\,.
\end{equation}
Note that the above is the Chamfer loss between the plane sample points \cite{smith2019geometrics}.

Finally, given a collection of planes $\mathcal{P}_1, \mathcal{P}_2, \ldots, \mathcal{P}_P$, the fitting loss, $\mathcal{L}_\text{task}$, is
\begin{equation}
    \mathcal{L}_\text{task}(\mathcal{M}) = \frac{1}{P}\sum_{i=1}^P L_\text{cf}(\mathcal{S}_{\mathcal{P}_i}(\mathcal{M}), \mathcal{T}_{\mathcal{P}_i})\,.
\end{equation}

\section{Fitting \ours{} to Voxel Segmentations}
To fit \ours{} to a binary 3D voxel image $\textbf{B} \in \R^{W\times H \times D}$ we first use the Euclidean distance transform to create a new image $\textbf{U} \in \R^{W\times H \times D}$ where each voxel contains the unsigned distance to the segmentation boundary. Given a mesh $\mathcal{M}$ with $V$ vertices, the fitting loss is then given by
\begin{equation}
    \mathcal{L}_\text{task}(\mathcal{M}) = \frac{1}{V}\sum_{v=1}^V \textbf{U}(\textbf{x}_v)\,,
\end{equation}
where $\textbf{U}(\textbf{x}_v)$ is trilinear interpolation of $\textbf{U}$ at vertex position $\textbf{x}_v$.

To optimize the latent vectors $\textbf{Z} \in \R^{V\times d}$ we require the gradient of \textbf{U} at $\textbf{x}_v$. To get a robust estimate, we use a Sobel operator to pre-compute a gradient image $\textbf{G} \in \R^{W\times H \times D \times 3}$ which contains the gradient of \textbf{U} at each voxel position. We then use trilinear interpolation to evaluate the gradient of \textbf{U} as $\nabla \textbf{U}(\textbf{x}_v) = \textbf{G}(\textbf{x}_v)$.

We also attempted to fit \ours{} directly to the binary segmentation or the softmax outputs of the CNN backbone. In practice, we found that the distance field gradients made it easier for the model to fit the images.

\end{document}